\documentclass[runningheads]{llncs}
\usepackage{todonotes}
\usepackage{graphicx}
\usepackage[utf8]{inputenc}
\usepackage{xcolor}
\usepackage{graphicx}
\usepackage{colortbl}
\usepackage{epstopdf}
\DeclareGraphicsExtensions{.png,.pdf}

\usepackage{subcaption}
\usepackage{comment}

\newcommand{\maple}{MapleLCMDistChronoBT}
\newcommand{\perm}{PERM}
\newcommand{\temp}{TEMP}

\begin{document}
\title{An Experimental Study of Permanently Stored Learned Clauses}
\author{Sima Jamali \and
David Mitchell }
\authorrunning{S. Jamali et al.}
\institute{Simon Fraser University, Vancouver, Canada \\
\email{\{sja88,dgm\}@sfu.ca}}

\maketitle              % typeset the header of the contribution
\begin{abstract}

Modern CDCL SAT solvers learn clauses rapidly, and an important 
heuristic is the clause deletion scheme.  Most current solvers 
have two (or more) stores of clauses. One has ``valuable'' 
clauses which are never deleted.  Most learned clauses are added 
to the other, with an aggressive deletion strategy to restrict 
its size.
Recent solvers in the MapleSAT family, have comparatively complex
deletion scheme, and perform well.  Many solvers store only 
binary clauses permanently, but \maple{} stores clauses with 
small LBD permanently.  We report an experimental study of 
the permanent clause store in \maple{}.
We observe that this store can get quite large, but several 
methods for limiting its size reduced performance.  
We also show that alternate size and LBD based criteria 
improve performance, while still having large permanent stores.
In particular, saving clauses up to size 8, and adding small 
numbers of high-centrality clauses, both improved performance, 
with the best improvement using both methods.

\end{abstract}
%
%
%%%%%%%%%%%%%%%%%%%%%%%%%%%%%%%%%%%%%%%%%%%%%
\section{Introduction}
%%%%%%%%%%%%%%%%%%%%%%%%%%%%%%%%%%%%%%%%%%%%%

High-performance CDCL SAT solvers often learn millions of clauses 
during a run.  Retaining all of these clauses permanently seems 
impractical, so most must be deleted. The scheme for deleting learned 
clauses is an important CDCL solver heuristic, usually based on 
measures of ``clause quality'' such as size or literal block 
distance (LBD) \cite{audemard2009predicting,ansotegui2015using} to estimate future utility.  
Most solvers store some ``high quality'' clauses permanently
and review the others periodically to delete some of lower 
quality.  We will call the set of clauses that are never deleted 
\perm{}, and the set which will be reviewed for deletion \temp{}.
(Some \perm{} clauses are deleted because they are satisfied 
by a learned unit clause, but these could not be used again anyway.)
\perm{} and \temp{} are often, but not always, stored in distinct data 
structures. 

We have performed many experiments on decision and deletion heuristics 
in solvers of the MapleSAT family \cite{liang2016maple,oh2016improving,liang2016learning,kochemazov2019maplelcmdistchronobt,nadel2018chronological,zhang2020relaxed}
, which have performed
very well in recent SAT solver competitions.
This experience convinced us that the clause deletion scheme is 
important to their performance, but the complexity of the 
scheme makes it hard to understand why.
Recent MapleSAT-based solvers have three distinct stores of clauses, 
use at least two dynamic clause quality measures (LBD and activity), 
and a number of heuristic rules to move clauses between the stores.  
In contrast, it is possible to build quite good solvers with much 
simpler schemes \cite{jamali2019simplifying}.  Many solvers place only binary clauses in \perm{} and 
use a simple measure to delete from \temp{}.  For example, 
Glucose and Cadical use LBD with ties broken by size \cite{audemard2009predicting,cadical2020solvers}.  
(Cadical also retains some \temp{} clauses based on two bits of 
recent use information.)

Here we report an empirical study of \perm{} in \maple{}.
We show that:
\begin{itemize}
   \item  Usually \perm{} is of moderate size, but sometimes it grows very large;
   \item At least some ways of restricting \perm{} size 
      for formulas where it gets large did not help.
   \item Alternate LBD and size based criteria for \perm{} can 
     improve performance (with similar-sized \perm{}).
     In particular (perhaps surprisingly) sending all 
     clauses of size up to 8 to \perm{} was very effective. 
   \item Adding a few very high-centrality (HC) clauses to 
     \perm{} improved performance on formulas for which 
      centrality computation is fast.  Our results 
      indicate that very high-centrality clauses are valuable
      even if they are long.
   \item The best improvement in our experiments comes 
     from a combination of size$\leq 8$ and adding 
     HC clauses to \perm{}. This version solved 197 instances, 
      13 more than the 184 that \maple{} solved.
   \item High-centrality clauses are used more often in conflict analysis.
   \item There are small clauses that are easy to derive, and 
    which help when added to \temp{}, but hurt when 
     added to \perm{}.
\end{itemize}

The base solver in all reported experiments is \maple{}, the 
first-place solver in the 2018 SAT solver competition~\cite{nadel2018chronological,heule2019sat}.
We denote simply ``maple'' in keys of some figures, to keep 
names short. 
The \maple{} deletion scheme was originally adopted from
COMiniSatPS~\cite{kochemazov2019maplelcmdistchronobt,oh2016improving}. 
Our data are for 400 formulas from the Main Track of the 2020 
competition with a 5000 second timeout \cite{sat2020competition}. Computations were performed on the Cedar compute cluster \cite{Cedar} 
operated by Compute Canada \cite{ComputeCanada}. The cluster consists 
of 32-core, 64 GB nodes with Intel “Broadwell” CPUs running at 2.1Ghz.
We use clause quality measures Size, LBD and Centrality.
Size is the number of literals in the clause. 
LBD~\cite{audemard2009predicting} is the number of  
decision levels of literals in the clause at the time it is computed
(at which time all literals must be assigned).
Clause Centrality \cite{sima2018structure} is the 
average betweenness centrality of its variables in the primal 
graph of the formula\cite{sima2017structure}.

%%%%%%%%%%%%%%%%%%%%%%%%%%%%%%%%%%%%%%%%%%%%%%%%%%
\section{Size and Value of \perm{} in \maple{}}
\label{sec:size-of-perm}
%%%%%%%%%%%%%%%%%%%%%%%%%%%%%%%%%%%%%%%%%%%%%%%%%%

We begin by considering the make-up and value of \perm{} in \maple{}. 
The clause database is partitioned into three sets called Core, Tier2 
and Local, stored in distinct databases.
Core stores the \perm{} clauses, and Tier2 and Local comprise \temp{}.
A new learned clause is sent to Core if $\mathrm{LBD} \leq 3$, 
to Local if $\mathrm{LBD}>6$, and to Tier2 otherwise.  If after 
the first 100,000 conflicts $|\mathrm{Core}|<100$, the core 
threshold is changed from $\leq 3$ to $\leq 5$.
The LBD of a clause is re-computed each time it is used in conflict analysis.
When the LBD of a clause is reduced, it is moved from Local to Tier2 or 
Core, or from Tier2 to Core, in accordance with the relevant thresholds.
Every 10,000 conflicts, any clause in Tier2 that has not been used 
for 30,000 conflicts is moved to Local.
Deletion is carried out only on clauses in Local, using a 
``Delete Half'' scheme based on a VSIDS-like clause activity 
measure \cite{oh2015between,oh2016improving}.

\vspace*{-0.5cm}
\begin{figure}[htp]
    \centering
    \hspace*{-0.8cm}
    \includegraphics[width=14cm]{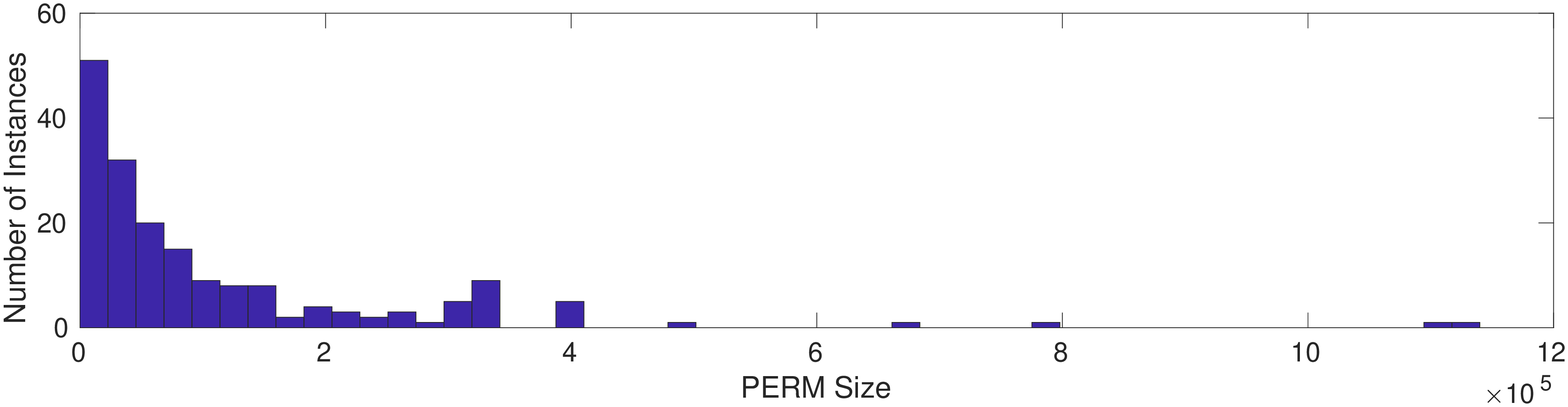}
    \caption{\perm{} Size Histogram}
    \label{fig:CoreDbHist0}
\end{figure}

\maple{} solved 184 instances from the benchmark set with the 5000 second 
time-out. 
Figure~\ref{fig:CoreDbHist0} is a histogram showing the number 
of clauses in \perm{} at the end of the run on these formulas.
For about half of the formulas the final size of \perm{} is 20,000 
or less, which is moderate compared to the average \temp{} size of 
about 30,000.  However for nearly a quarter of the instances (22\%) 
the final \perm{} size is more than 150,000.  

\begin{figure}[htb]
    \centering
    \hspace{-0.90cm}
    \includegraphics[width=12cm]{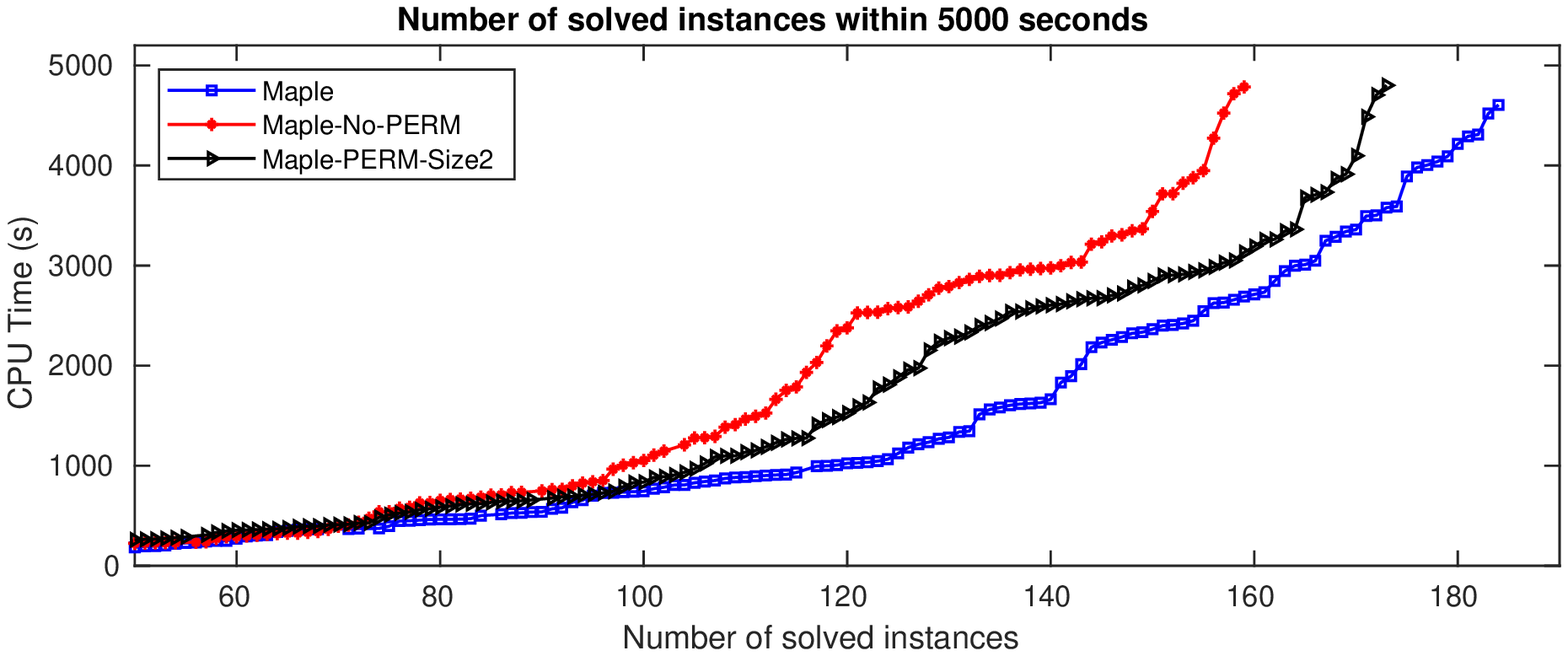}
    \caption{Effect of removing \perm{} from \maple{}}
    \label{fig:cactusNoCore0}
\end{figure}

\begin{figure}[tb]
    \centering
    \hspace{-0.95cm}
    \includegraphics[width=12.5cm]{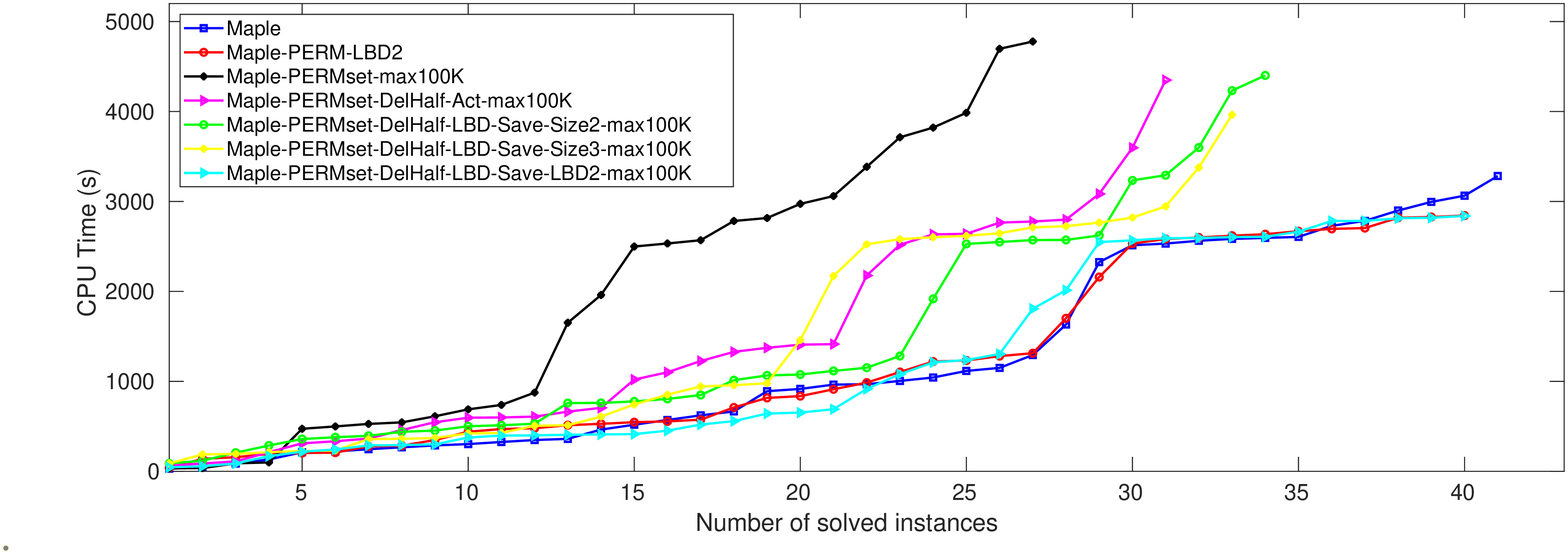}
    \caption{Effect of limiting size of \perm{} on ``LC'' formulas.}
    \label{fig:cactusDelCore}
\end{figure}

We report two experiments to evaluate the usefulness of \perm{}
in \maple{}. The first compares the default solver 
with two modified versions, one with \perm{} empty, and one with 
only binary clauses sent to \perm{}.
Figure~\ref{fig:cactusNoCore0} shows that
both modifications reduce performance substantially.  
Conventional wisdom suggests 150,000 is very large for a learned 
clause set, and we hypothesized it might be better to limit its size.
In the second experiment, we use various schemes to restrict 
the size of \perm{}, with the goal of keeping it less than 100,000. 
We applied these schemes to the formulas for which \perm{} grew 
to more than 150,000 clauses. 
As Figure~\ref{fig:cactusDelCore} shows, even for these formulas 
most of the methods were damaging to performance, and even the 
best do not seem beneficial.  
The schemes are as follows:
\begin{itemize}
\item Maple-PERM-LBD2: Require LBD $\leq 2$.
\item Maple-PERMset-max100K: 
If size of \perm{} reaches 100,000, send no more clauses to \perm{} and 
send all new clauses to \temp{}.
\item Maple-PERMset-DelHalf-Act-max100K: 
If size of \perm{} reaches 100,000, invoke a ``delete-half'' deletion scheme 
on \perm{}, based on clause activity.
\item Maple-PERMset-DelHalf-LBD-Save-X-max100K: 
If size of \perm{} reaches 100,000, invoke a ``delete-half'' deletion scheme 
on \perm{}, based on LBD (with ties broken by clause age), but never deleting 
clauses with property X, for X in 
$\{ \mathrm{size}\leq 2, \mathrm{size}\leq 3, \mathrm{LBD}\leq 2 \}$.
\end{itemize}

%\vspace*{-0.9cm}
\section{Varying Size and LBD Criteria for \perm{}}
\label{sec:Varying}

\begin{figure}[htp]
\includegraphics[width=12cm]{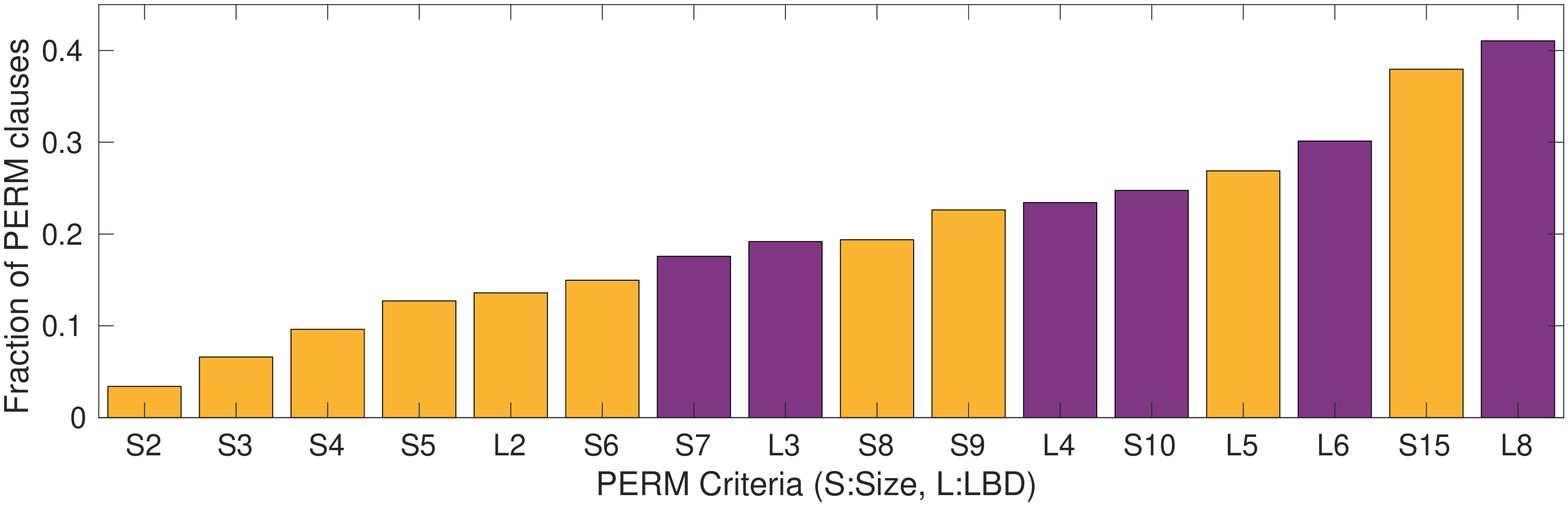} 
\includegraphics[width=12cm]{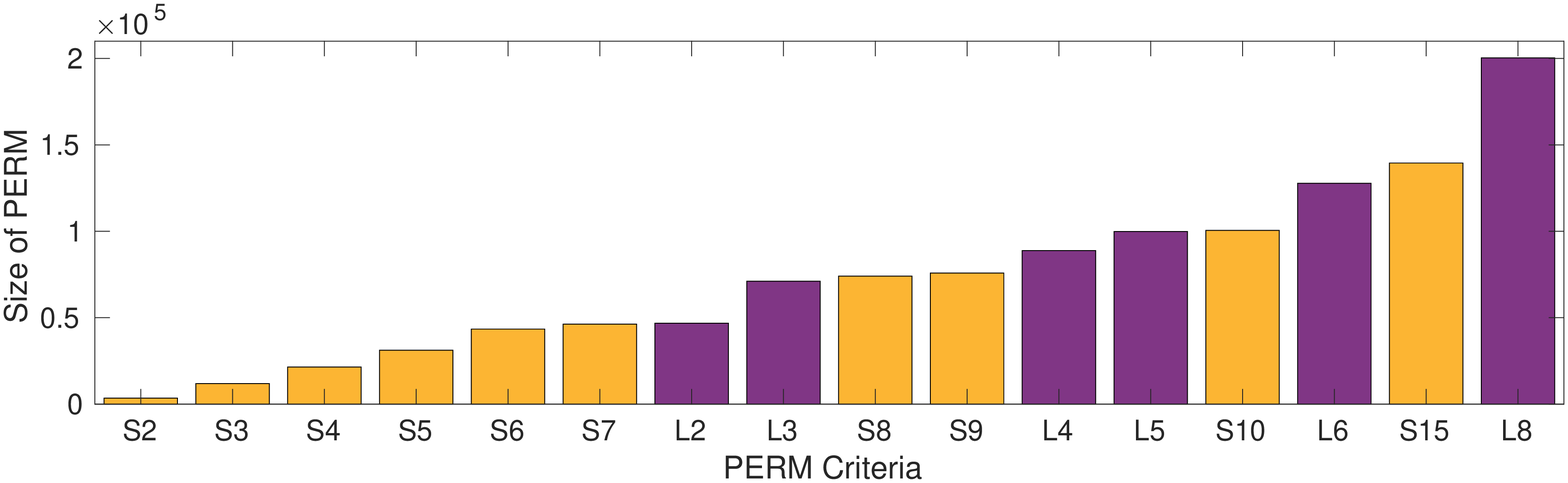} 
\caption{Fraction of learned clauses sent to \perm{} (upper), 
and final size of \perm{} (lower) with varied \perm{} criteria.}
\label{fig:diffCore}
\end{figure}

%\vspace*{-0.1cm}
The \maple{} criterion for putting in \perm{} is 
LBD $\leq 3$. (Sometimes it is changed to LBD $\leq 5$ 
during the run but for few formulas.)
We report an experiment in which we vary the criteria 
over two ranges:  Size $\leq k$, for $k\in\{2,\ldots 15\}$
and LBD $\leq k$, for $k \in \{2,\ldots 8\}$.
%%Figure~\ref{fig:CoreRate2} shows the effect on the fraction 
Figure~\ref{fig:diffCore} shows the effect on the fraction 
of learned clauses sent to \perm{}, and on the final 
size of \perm{}. 

\begin{figure}[htp]
\includegraphics[width=12cm]{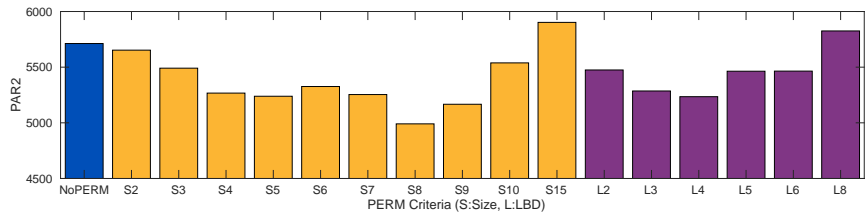} 
\caption{Effect of \perm{} criterion on Par-2 Scores}
\label{fig:par2-diffcore11}
\end{figure}

\begin{figure}[htp]
    \centering
    \includegraphics[width=12cm]{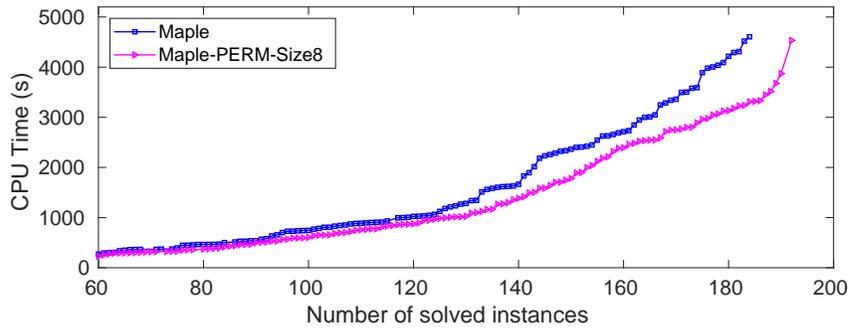}
    \caption{Performance of solvers with small clauses in Core}
    \label{fig:cactusSizeCore}
\end{figure}

Figure~\ref{fig:par2-diffcore11} shows the effect of these 
variations on Par-2 performance scores \cite{SATwebsite}.  
The best Par-2 score is with Size $\leq8$.
%%As Figure~\ref{fig:DBsize-diffCore22} shows, the number of 
As Figure~\ref{fig:diffCore} shows, the number of 
clauses saved to \perm{} with Size $\leq 8$ falls between 
that for LBD $\leq 3$ and LBD $\leq 4$, and is only 
slightly larger than for the default version.
Figure~\ref{fig:cactusSizeCore} compares the performance 
of the Size $\leq 8$ version with the default LBD $\leq 3$ 
in a cactus plot.

%\begin{figure}[htp]
%\includegraphics[width=12cm]{figures/Core_rate2.eps} 
%\caption{Fraction of learned clauses sent to \perm{}, with various \perm{} criteria.}
%\label{fig:CoreRate2}
%\end{figure}

%\begin{figure}[htp]
%\includegraphics[width=12cm]{figures/Core_size_end2.eps} 
%\caption{Final size of \perm{} with various \perm{} criteria.}
%\label{fig:DBsize-diffCore22}
%\end{figure}

%%%%%%%%%%%%%%%%%%%%%%%%%%%%%%%%%%%%%%%%%%%%%%%%%%%%%%%%%%
\section{Adding High-Centrality Clauses to \perm{}}

Clause betweenness centrality as defined in \cite{sima2018structure} 
has shown to be a useful clause quality measure.  
Here we add a limited number of high-centrality (HC) clauses 
to \perm{} in \maple{}. We computed variable centralities using 
the Brandes algorithm \cite{brandes2001faster} in the solver.
The centrality computation sometimes takes too long, and we 
limited it to 150 seconds, obtaining centralities for 168 of 
the 400 instances.  For the other formulas we did not use 
centrality.
We normalize the centrality values by $1/(n-1)(n-2)$ where $n$ is the number of variables, so they fall in $[0,1]$. 
The HC clauses can be large and result in computation and memory overhead so care is needed when adding them to \perm{}. We aimed to include at least the 0.02\% of learned clauses with highest centrality.
We set an initial centrality threshold of $CT \geq 0.008$ which was chosen empirically.
Every 100,000 conflicts, if the number HC clauses in \perm{} is 
less than 0.02\% of all learned clauses, $CT$ is reduced by 0.001, 
but it is never reduced below 0.001. 
We report three versions:
%%\vspace*{-0.3cm}
\begin{itemize}
    \item Maple-PERM-HC-max10K: Add at most the first 10K HC 
    clauses to \perm{}.
    \item Maple-PERM-HC-max25K: Add at most the first 25K HC 
    clauses to \perm{}.
    \item Maple-PERM-HC-Size15-max10K: Add HC clauses to \perm{} 
    only if they have size $\leq 15$, adding at most the first 10K.
\end{itemize}

\begin{figure}[htp]
    \centering
    \includegraphics[width=12cm, height=3.5cm]{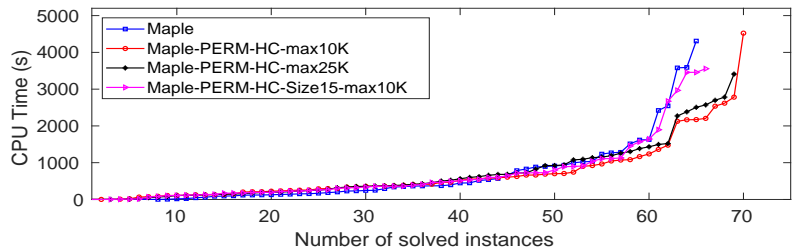}
    \caption{Performance of with high-centrality clauses in \perm}
    \label{fig:cactusCent}
\end{figure}

Figure \ref{fig:cactusCent} compares the performance 
of these versions, including centrality computations, against the default.  The two version 
with no size limit on the HC clauses performed 
noticeably better.  The version with the size limit 
performed only slightly better than the default.
This indicates long HC clauses are valuable.
The average number of additional HC \perm{} clauses (that would not have been placed in \perm{} because of LBD) was 8,200 in the version with the limit of 10K, 16,000 with the limit of 25K, and 7050 with the limit of 
10K and size $\leq 15$.

\begin{figure}[htp]
    \centering
    \includegraphics[width=12cm, height=4cm]{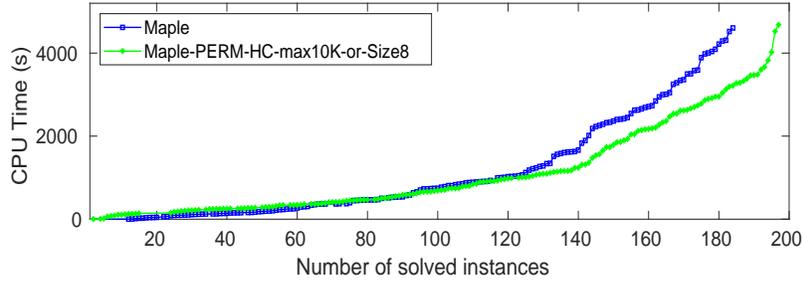}
    \caption{Performance with \perm{} criteria Size$\leq 8$ or HC.}
    \label{fig:cactusCentSize8}
\end{figure}

Often the combination of two heuristics that are beneficial 
does not improve over the best of the two.  However, in 
the case of our criteria for \perm{}, the following 
combination did improve overall performance: For each 
instance, if did not get centralities within the 150 
seconds time limit, we used \perm{} criterion of Size$\leq 8$; 
otherwise we used LBD$\leq 3$ and added HC clauses.
Figure~\ref{fig:cactusCentSize8} compares this 
version with the original. It solves 13 more instances.

\vspace*{-0.5cm}
%%\vspace*{-1.55cm}

\begin{figure}[htp]
\includegraphics[width=12cm, height=3cm]{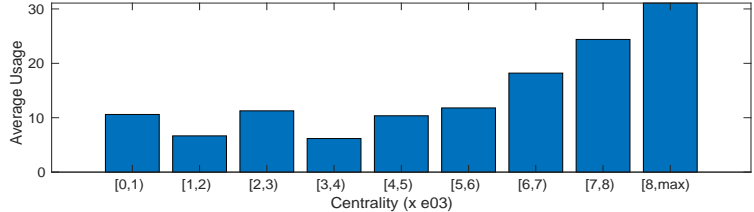}
\caption{Average Usage with respect to Centrality of clauses}
\label{fig:usage-cent}
\end{figure}

Long clauses are generally less valuable than short clauses.
They eliminate fewer truth assignments, and are used less 
frequently in CDCL solvers.  However, our data show that 
the value of HC clauses added to \perm{} comes from the 
longer ones. Thus, we would like to understand the 
usage of HC clauses better.   
We define the Clause Usage to be the number of times 
a clause is used during conflict analysis.   We ran \maple{} 
with clause deletion turned off, and stopped the execution 
after 500,000 conflicts (so 500,000 clause have been learned).
Then looked at clause usage as a function of centrality.
Figure~\ref{fig:usage-cent} is a histogram of the usage rates 
for all learned clauses obtained from 10 formulas randomly
chosen from our benchmark.  The figure shows clearly that 
HC clauses were used more than others, suggesting they may 
be more useful in generating new conflicts.

%%%%%%%%%%%%%%%%%%%%%%%%%%%%%%%%%%%%%%%%%%%%%%%%%%%%%%
\section{Small good clauses not to add to \perm{}}
%%%%%%%%%%%%%%%%%%%%%%%%%%%%%%%%%%%%%%%%%%%%%%%%%%%%%%

In Section~\ref{sec:Varying}, we showed that performance 
improved when we saved all learned clauses of size up to 8 
to \perm{}.  Here, we show that there are small clauses 
we can derive simply which help when added to \temp{} 
but not when added to \perm{}.
Standard conflict analysis schemes derive one clause, called 
the 1-UIP (for First Unique Implication Point) clause, at 
each conflict.  Various other schemes have been tried, but 
most reports confirm that the 1-UIP scheme is 
best~\cite{zhang2001efficient,dershowitz2007towards,sorensson2009minimizing}. 
An example of adding more clauses 
is in \cite{dershowitz2007towards}, but these clauses 
require significant additional reasoning.  

Here, we introduce a simple scheme to learn additional small 
clauses which are very cheap to obtain but still improve 
performance.  Regarding the focus of this paper, 
the interesting observation is that they have length less 
than 8, but adding them to \perm{} reduces performance 
while adding them to \temp{} improves performance. 
(In contrast, adding all small 1-UIP clauses to \perm{} 
improves performance, as we showed above.)

\begin{figure}[htp]
    \centering
    \includegraphics[width=12cm]{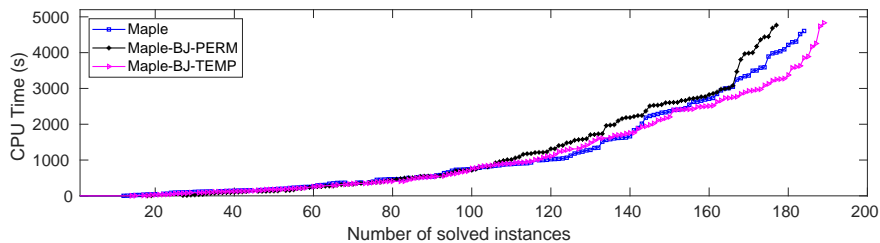}
    \caption{Performance of solvers with new learned clauses}
    \label{fig:cactusSizeCoreBJ0}
\end{figure}

Assume a conflict at level $x$, meaning after assigning $x$ 
literals $l_1, l_2, .., l_x$to true, a conflict is reached. 
After conflict analysis the solver backjumps to a level $b$ and 
learns a 1UIP clause $C = \{ m_1, m_2,... , m_{i-1}, m_i\}$. 
Only one literal $m_i$ from $C$ belongs to level $x$, 
and $b<x$, so after the first $b$ decisions, if we had $C$ 
in the clause database unit propagation we would prevent 
this conflict by assigning $m_i$ true.  Therefore, we 
can also learn clause 
$C_2 = \{ \neg l_1, \neg l_2,... \neg l_b, m_i \}$. 
If $b<6$, this clause has size $<=6$, so we have a new 
small clause with little work.
Here the last two literals of $C_2$, $\neg l_b$ and $m_i$, 
are glued together so $C_2$ has LBD $|b|$ and size $|b|+1$. 
We added clauses of this kind and added them either to 
\perm{} or \temp{}.  
Figure~\ref{fig:cactusSizeCoreBJ0} compares performance of 
these version with \maple{}. In the formulas of this experiment, an average of 8500 additional clauses were generated using this scheme most of which had LBD 4 or 5. This can be a factor in making them less interesting for \perm{}.

\vspace*{-0.2cm}

%%%%%%%%%%%%%%%%%%%%%%%%%%%%%%%%%%%%%%%%%%%
\vspace*{-0.9cm}
\section{Discussion}

We have performed a number of experiments on criteria for 
learned clauses in \maple{} to be saved permanently 
(the \perm{} set). 
Our experiments confirm that even a large \perm{} set 
helps performance, using several criteria related to 
size and LBD, which are widely used as clause quality measures.   
We also showed that adding a small number of very 
high centrality clauses, when centralities could be computed, 
improved performance.  A solver version in which the  
\perm{} criteria was either size $\leq 8$ or LBD$\leq 3$ 
plus HC clauses, depending on availability of centrality 
information, solved 13 more formulas than the original 
\maple{} on a benchmark of 400 competition instances. 

%%%%%%%%%%%%%%%%%%%%%%%%%%%%%%%%%%%%%%%%%%%%%%%%%%%%%%%%%%%%%

%%%%%%%%%%%%%%%%%%%%%%%%%%%%%%%%%%%%%%%%

%%%%%%%%%%%%%%%%%%%%%%%%%%%%%%%%%%%%

\bibliographystyle{plain}
\bibliography{references}

\begin{thebibliography}{10}

\bibitem{SATwebsite}
Sat competitions : http://www.satcompetition.org/.

\bibitem{ansotegui2015using}
Carlos Ans{\'o}tegui, Jes{\'u}s Gir{\'a}ldez-Cru, Jordi Levy, and Laurent
  Simon.
\newblock Using community structure to detect relevant learnt clauses.
\newblock In {\em International Conference on Theory and Applications of
  Satisfiability Testing}, pages 238--254. Springer, 2015.

\bibitem{audemard2009predicting}
Gilles Audemard and Laurent Simon.
\newblock Predicting learnt clauses quality in modern sat solvers.
\newblock In {\em Twenty-first International Joint Conference on Artificial
  Intelligence}, 2009.

\bibitem{sat2020competition}
Tomáš Balyo, Nils Froleyks, Marijn~J.H. Heule, Markus Iser, Matti Järvisalo,
  and Martin Suda.
\newblock Proceedings of sat competition 2020 : Solver and benchmark
  descriptions.
\newblock 2020.

\bibitem{cadical2020solvers}
Armin Biere, Katalin Fazekas, Mathias Fleury, and Maximillian Heisinger.
\newblock {CaDiCaL}, {Kissat}, {Paracooba}, {Plingeling} and {Treengeling}
  entering the {SAT Competition 2020}.
\newblock In {\em Proc.~of {SAT Competition} 2020 -- Solver and Benchmark
  Descriptions}, volume B-2020-1 of {\em Department of Computer Science Report
  Series B}, pages 51--53. University of Helsinki, 2020.

\bibitem{brandes2001faster}
Ulrik Brandes.
\newblock A faster algorithm for betweenness centrality.
\newblock {\em Journal of mathematical sociology}, 25(2):163--177, 2001.

\bibitem{Cedar}
{Cedar}, {A Compute Canada Cluster}.
\newblock \url{https://docs.computecanada.ca/wiki/Cedar}.

\bibitem{ComputeCanada}
{Compute Canada}: {Advanced Research Computing {(ARC)} Systems}.
\newblock \url{https://www.computecanada.ca/}.

\bibitem{dershowitz2007towards}
Nachum Dershowitz, Ziyad Hanna, and Alexander Nadel.
\newblock Towards a better understanding of the functionality of a
  conflict-driven sat solver.
\newblock In {\em International Conference on Theory and Applications of
  Satisfiability Testing}, pages 287--293. Springer, 2007.

\bibitem{heule2019sat}
Marijn~JH Heule, Matti J{\"a}rvisalo, and Martin Suda.
\newblock Sat competition 2018.
\newblock {\em Journal on Satisfiability, Boolean Modeling and Computation},
  11(1):133--154, 2019.

\bibitem{sima2017structure}
Sima Jamali and David Mitchell.
\newblock Improving sat solver performance with structure-based preferential
  bumping.
\newblock In {\em Global Conference of Artificial Intelligence (GCAI)}, pages
  175--187. Springer, 2017.

\bibitem{sima2018structure}
Sima Jamali and David Mitchell.
\newblock Centrality-based improvements to cdcl heuristics.
\newblock In {\em International Conference on Theory and Applications of
  Satisfiability Testing}, pages 122--131. Springer, 2018.

\bibitem{jamali2019simplifying}
Sima Jamali and David Mitchell.
\newblock Simplifying cdcl clause database reduction.
\newblock In {\em International Conference on Theory and Applications of
  Satisfiability Testing}, pages 183--192. Springer, 2019.

\bibitem{kochemazov2019maplelcmdistchronobt}
Stepan Kochemazov, Oleg Zaikin, Victor Kondratiev, and Alexander Semenov.
\newblock Maplelcmdistchronobt-dl, duplicate learnts heuristic-aided solvers at
  the sat race 2019.
\newblock {\em Proceedings of SAT Race}, pages 24--24, 2019.

\bibitem{liang2016learning}
Jia~Hui Liang, Vijay Ganesh, Pascal Poupart, and Krzysztof Czarnecki.
\newblock Learning rate based branching heuristic for sat solvers.
\newblock In {\em International Conference on Theory and Applications of
  Satisfiability Testing}, pages 123--140. Springer, 2016.

\bibitem{liang2016maple}
Jia~Hui Liang, Chanseok Oh, Vijay Ganesh, Krzysztof Czarnecki, and Pascal
  Poupart.
\newblock Maple-comsps, maplecomsps lrb, maplecomsps chb.
\newblock {\em SAT Competition}, page~52, 2016.

\bibitem{nadel2018chronological}
Alexander Nadel and Vadim Ryvchin.
\newblock Chronological backtracking.
\newblock In {\em International Conference on Theory and Applications of
  Satisfiability Testing}, pages 111--121. Springer, 2018.

\bibitem{oh2015between}
Chanseok Oh.
\newblock Between sat and unsat: the fundamental difference in cdcl sat.
\newblock In {\em International Conference on Theory and Applications of
  Satisfiability Testing}, pages 307--323. Springer, 2015.

\bibitem{oh2016improving}
Chanseok Oh.
\newblock {\em Improving SAT solvers by exploiting empirical characteristics of
  CDCL}.
\newblock PhD thesis, New York University, 2016.

\bibitem{sorensson2009minimizing}
Niklas S{\"o}rensson and Armin Biere.
\newblock Minimizing learned clauses.
\newblock In {\em International Conference on Theory and Applications of
  Satisfiability Testing}, pages 237--243. Springer, 2009.

\bibitem{zhang2001efficient}
Lintao Zhang, Conor~F Madigan, Matthew~H Moskewicz, and Sharad Malik.
\newblock Efficient conflict driven learning in a boolean satisfiability
  solver.
\newblock In {\em IEEE/ACM International Conference on Computer Aided Design.
  ICCAD 2001. IEEE/ACM Digest of Technical Papers (Cat. No. 01CH37281)}, pages
  279--285. IEEE, 2001.

\bibitem{zhang2020relaxed}
Xindi Zhang and Shaowei Cai.
\newblock Relaxed backtracking with rephasing.
\newblock {\em SAT COMPETITION 2020}, page~15.

\end{thebibliography}

\end{document}